\documentclass[10pt,twocolumn,letterpaper]{article}

\usepackage{colortbl}
\usepackage{cvpr}
\usepackage{times}
\usepackage{epsfig}
\usepackage{graphicx}
\usepackage{amsmath}
\usepackage{amssymb}

\usepackage{microtype}

\usepackage{booktabs}
\usepackage{threeparttable}
\usepackage[table]{xcolor}
\usepackage{multirow}
\usepackage{subfigure}
\usepackage{soul}

\usepackage[breaklinks=true,letterpaper=true,colorlinks,bookmarks=false]{hyperref}

\cvprfinalcopy 

\def\cvprPaperID{?} 

\ifcvprfinal\fi

\definecolor{yelloworange}{RGB}{255, 153, 0}
\definecolor{ultramarineblue}{RGB}{65, 102, 245}

\newtheorem{lemma-ap}{Lemma}

\newtheorem{claim-ap}{Claim}

\makeatletter
\usepackage{xspace}
\def\@onedot{\ifx\@let@token.\else.\null\fi\xspace}
\DeclareRobustCommand\onedot{\futurelet\@let@token\@onedot}

\newcommand{\eat}[1]{}

\makeatother

\begin{document}

\title{Multi-Scale Fully Convolutional Network for Cardiac Left Ventricle Segmentation}

\author{Han Kang$^{1,2}$ Defeng Chen$^{1,2}$\\
$^1$The School of Mathematical Science, Capital Normal University, Beijing\\
$^2$Beijing Advanced Innovation Center for Imaging Technology, Beijing\\
{\tt\small 2150502137@cnu.edu.cn, chendf@cnu.edu.cn}
}

\maketitle

\begin{abstract}
The morphological structure of left ventricle segmented from cardiac magnetic resonance images can be used to calculate key clinical parameters, and it is of great significance to the accurate and efficient diagnosis of cardiovascular diseases. Compared with traditional methods, the segmentation algorithms based on fully convolutional neural network greatly improve the accuracy of semantic segmentation. For the problem of left ventricular segmentation, a new fully convolutional neural network structure named MS-FCN is proposed in this paper. The MS-FCN network employs a multi-scale pooling module to ensure that the network maximises the feature extraction ability and uses a dense connectivity decoder to refine the boundaries of the object. Based on the Sunnybrook cine-MR dataset provided by the MICCAI 2009 challenge, numerical experiments demonstrate that our proposed model has obtained state-of-the-art segmentation results: the Dice score of our method reaches 0.93 on the endocardium, and 0.96 on the epicardium.

~\\

\end{abstract}

\section{Introduction}
Cardiac cine magnetic resonance imaging (MRI) is identified as the gold standard for the diagnosis and treatment of cardiovascular diseases. The contours of the left ventricle (LV) extracted from card
iac MRI can be used to calculate clinical parameters such as ventricular volume, myocardial mass, cardiac output, ejection fraction, and wall thickness. Due to the complexity of heart structure and th
e large amount of cardiac MRI data, the traditional manual delineation is inefficient and labor-intensive. Therefore, methods for LV endocardium (Endo) and epicardium (Epi) segmentation are always the
 research hotspots in the field of cardiac MR image analysis.

A considerable number of researches on fully automated or semi-automated segmentation methods for LV MR images have been proposed. The traditional methods can be divided roughly into: 1) segmentation
methods based on curve evolution, such as active contours \cite{Makowski2002}; 2) segmentation methods based on statistical shape model, such as active appearance model (AAM) \cite{Andreopoulos2008},
active shape model (ASM) \cite{Santiago2016}; 3) segmentation methods based on graph theory, such as graph cut method \cite{Lin2005}; 4) segmentation methods based on the combination of neural network
s and the former methods \cite{Avendi2016, Ngo2017}.

In recent years, driven by powerful deep neural networks, convolutional neural networks (CNNs) have achieved remarkable results in the field of semantic segmentation \cite{Luo2017, Chen2018}, represented by Fully Convolutional Network(FCN) \cite{Long2015}. FCN discards all fully connected layers in classification architecture and applies a fully convolutional structure to realize pixel-wise prediction. Later, researchers propose symmetrical encoder-decoder network structure, in which the encoder uses pooling layers to gradually reduce the spatial resolution of the input image, while the decoder gradually recovers the details of the target and correspondingly spatial resolution. U-net \cite{Ronneberger2015}, SegNet \cite{Badrinarayanan2015} and FC-DenseNet \cite{Jegou2016} are representatives of this network structure.

CNNs-based segmentation algorithms are also applied to the left ventricular segmentation and achieve great segmentation effects. Tran \cite{Tran2016} is the first person who apply FCN to cardiac MRI images segmentation and achieves a start-of-the-art result beyond previous automated methods. Romaguera \cite{Romaguera2017} trains FCN model by using whole cardiac images as input without cropping and obtains a good performance for LV segmentation. Khened \cite{Khened2018} imports modified FC-DenseNet \cite{Jegou2016} architecture for cardiac segmentation, which is parameter and memory efficient compared with FCN. They add parallel pathways in the initial convolution layer, and introduce short-cut connections in the upsampling path. Tan \cite{Tan2018} designs three separate neural networks, which are used to locate the left ventricle, estimate the LV centerpoint, and compute the radial distances of the endocardium and epicardium, respectively.

In this paper, we propose a new CNN architecture for cardiac LV segmentation which adopts a conventional encoder-decoder structure and we call it as Multi-Scale Fully Convolutional Network (MS-FCN). A multi-scale pooling module is employed in the encoding stage while a dense connection structure is used in the decoding stage, and the effects of them are all explained in detail in the following sections. Moreover, we conduct a series of ablation experiments on the network structure. Finally, we demonstrate state-of-the-art results on the public Sunnybrook dataset, which comes from the MICCAI 2009 Cardiac MR Left Ventricle Segmentation Challenge \cite{Radau2009}.

\section{Network architecture}
\label{sec:network}
In this section, we briefly introduce our neural network architecture and describe the concept of network design.

\begin{figure*}[!t]
  \centering
  \scalebox{0.8}{
  \begin{tabular}{c}
    \includegraphics[width=1\linewidth]{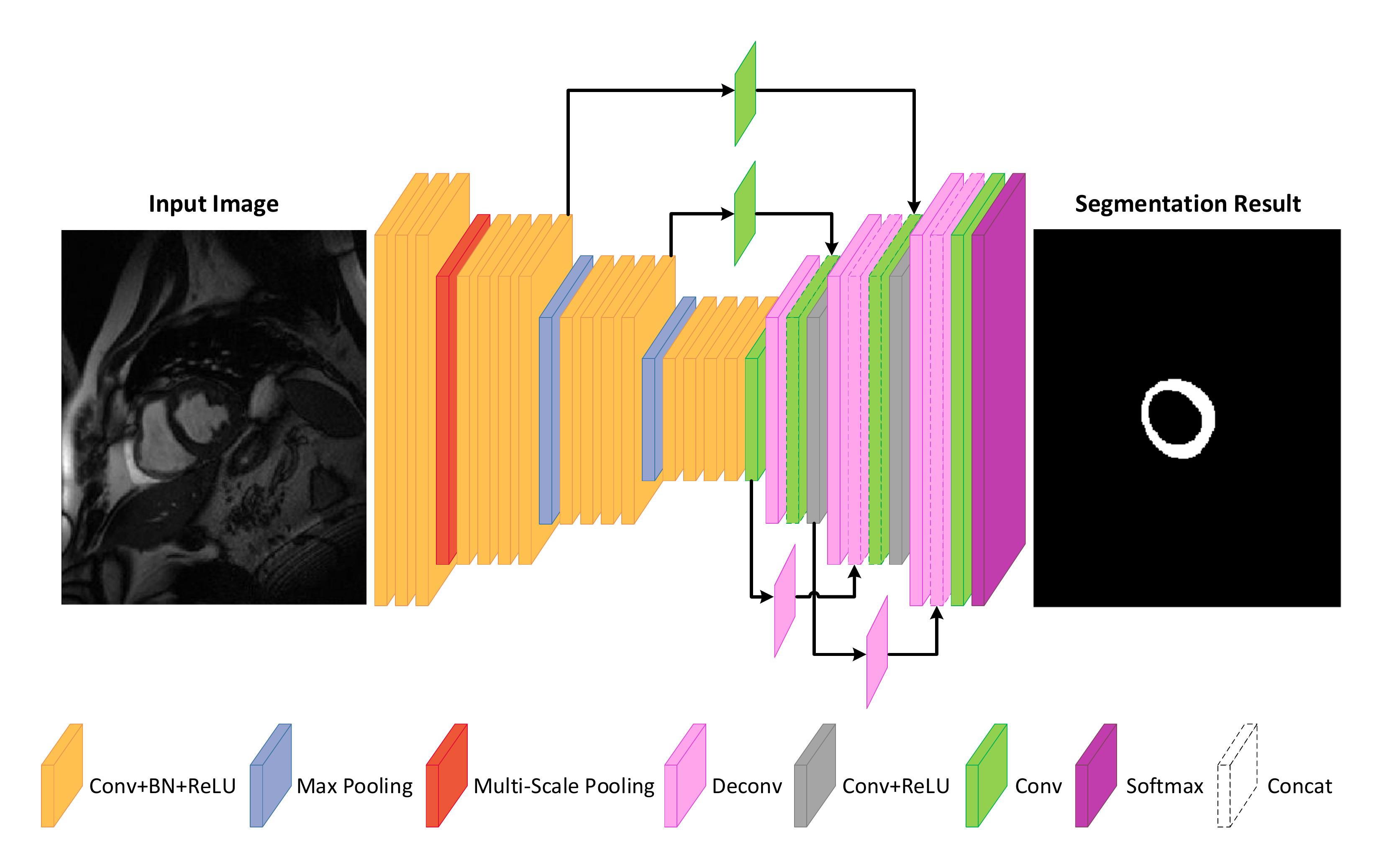} \\
  \end{tabular}
  }
  \caption{Overview of our proposed MS-FCN. In the encoder, we apply a multi-scale pooling module to obtain contextual information, while a dense connection structure is used to recover the object boundaries in the decoder.}
  \label{fig:model}
\end{figure*}

Figure \ref{fig:model} illustrates our proposed MS-FCN architecture which adopts an encoder-decoder structure. In the encoding stage, MS-FCN comprises 15 $3 \times 3$ convolution layers and three downsampling operations. Each convolution layer is followed by a Batch Normalization (BN) operation and a Rectified Linear Unit (ReLU) activation function. Unlike the first downsampling layer of FCN \cite{Tran2016}, MS-FCN employs a multi-scale pooling module instead of a conventional pooling layer to obtain more contextual information. In the decoding stage, in order to restore high-resolution detail, we use a cascade structure called dense connection structure. In addition, MS-FCN also uses skip connections similar to U-net \cite{Ronneberger2015}, which concatenate the feature maps of the encoding stage with the corresponding feature maps of the decoding stage. This is beneficial for the forward flow of feature maps and the backward propagation of gradient information.

To thoroughly understand the network structure, the multi-scale pooling module and the dense connection structure are introduced in detail.

\subsection{Multi-scale pooling module}
\label{subsec:encoder}
Inspired by PSPNet's pyramid pooling module \cite{Zhao2016}, we propose a hierarchical structure called multi-scale pooling module, which is used to maximises the feature extraction ability. The main difference from PSPNet is that the upsampling process inside the multi-scale pooling module is implemented by deconvolution but not bilinear interpolation. Compared to using fixed bilinear interpolation for upsampling, using a deconvolution layer with learnable parameters is better to restore rich image details, as demonstrated in Section \ref{subsec:upsampling_ms}.

\begin{figure}[!t]
  \centering
  \scalebox{1}{
  \begin{tabular}{c}
    \includegraphics[width=1\linewidth]{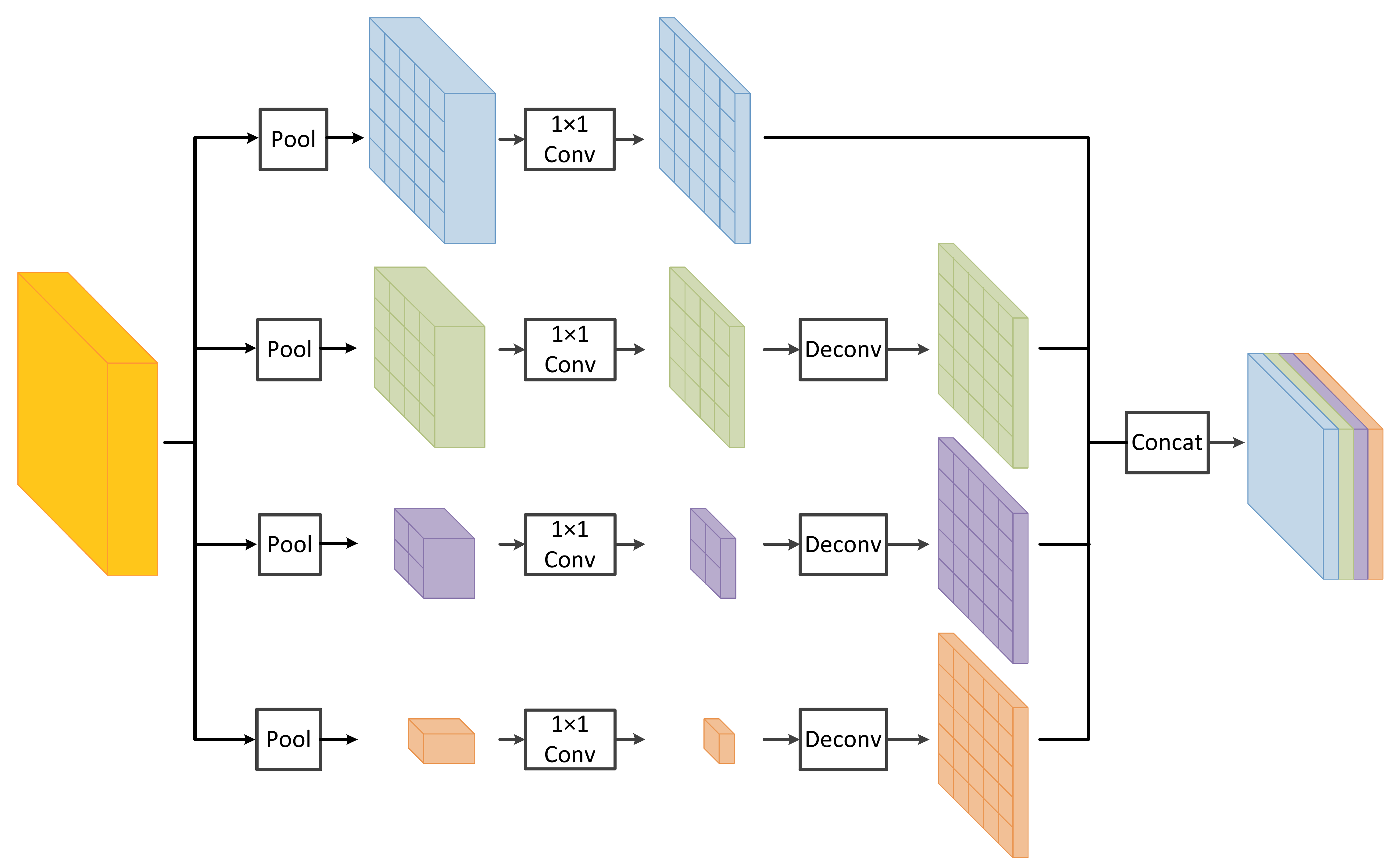} \\
  \end{tabular}
  }
  \caption{Multi-scale pooling module which includes four subpaths in parallel.}
  \label{fig:ms_pooling}
\end{figure}

The multi-scale pooling module consists of multiple subpaths in parallel, as illustrated in Figure \ref{fig:ms_pooling}. The first subpath uses a pooling layer with ratio $k$ as a baseline. The following subpaths employ pooling layers of different ratios which are all greater than $k$. Then, we employ a $1 \times 1$ convolutional layer in each path as the compression layer to reduce the number of feature maps. After that, we use deconvolution layers to upsample the low-dimension feature maps to the same size as the baseline. Finally, features of all subpaths are concatenated as the multi-scale output features.

It is noteworthy that the number of subpaths and the scaling ratio of the pooling layer in each subpath are all adjustable in the multi-scale pooling module, and they mainly depend on the input image resolution. Thus the downsampling ratios should maintain reasonable intervals to abstract additional contextual information. The multi-scale pooling module in MS-FCN includes four subpaths in parallel, where the downsampling ratios in the four subpaths are $2 \times 2$, $6 \times 6$, $18 \times 18$, and $36 \times 36$, respectively.

\subsection{Dense connection structure}
\label{subsec:decoder}
\begin{figure}
  \centering
  \subfigure[Gradually upsampling]{
  \begin{minipage}{0.95\linewidth}
    \centering
    \includegraphics[width=1\linewidth]{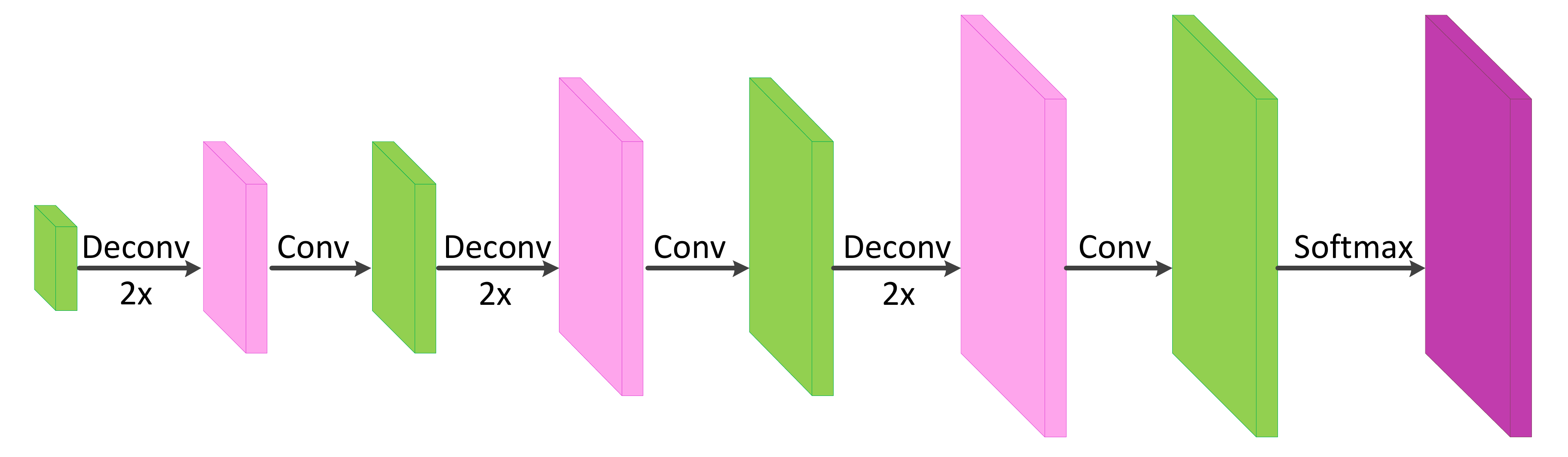}
    \label{fig:decoder1}
  \end{minipage}
  }
  \subfigure[Dense connection upsampling]{
  \begin{minipage}{0.95\linewidth}
    \includegraphics[width=1\linewidth]{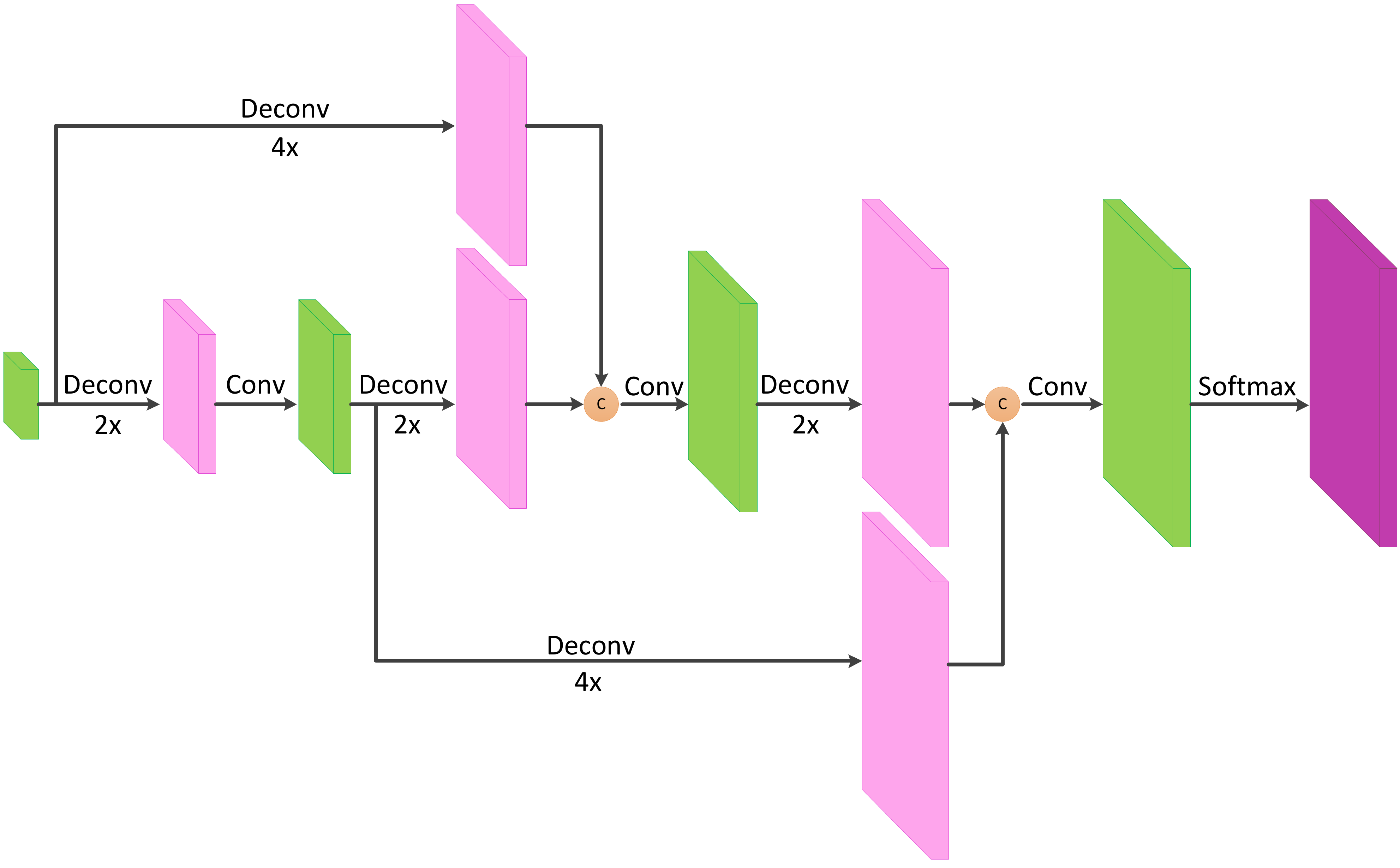}
    \label{fig:decoder2}
  \end{minipage}
  }
  \caption{Different upsampling strategies used in the decoder. Usual decoding stage employs upsampling layers and convolution layers alternately(a), while our network employs the dense connection upsampling(b).}
  \label{fig:decoder}
\end{figure}

In the decoding stage, most semantic segmentation networks \cite{Badrinarayanan2015, Chen2017DeepLabv2, Fu2017} usually employ upsampling layers and convolution layers alternately for decoding, as shown in Figure \ref{fig:decoder1}. Because the upsampling multiple in each step is fixed, the deconvolution layers can only extract limited features information from the decoded image.

In view of this phenomenon, we propose a dense connection structure which is illustrated in Figure \ref{fig:decoder2}. In this structure, low resolution feature maps are upsampled in different multiples, and then the upsampled feature maps with the same resolution are concatenated, later the subsequent convolution operations are performed. In this way, the information of low-resolution feature maps can flow through different paths in the network, thereby increasing the possibility of refining the LV boundaries.

\begin{figure}[!t]
  \centering
  \scalebox{1}{
  \begin{tabular}{c}
    \includegraphics[width=1\linewidth]{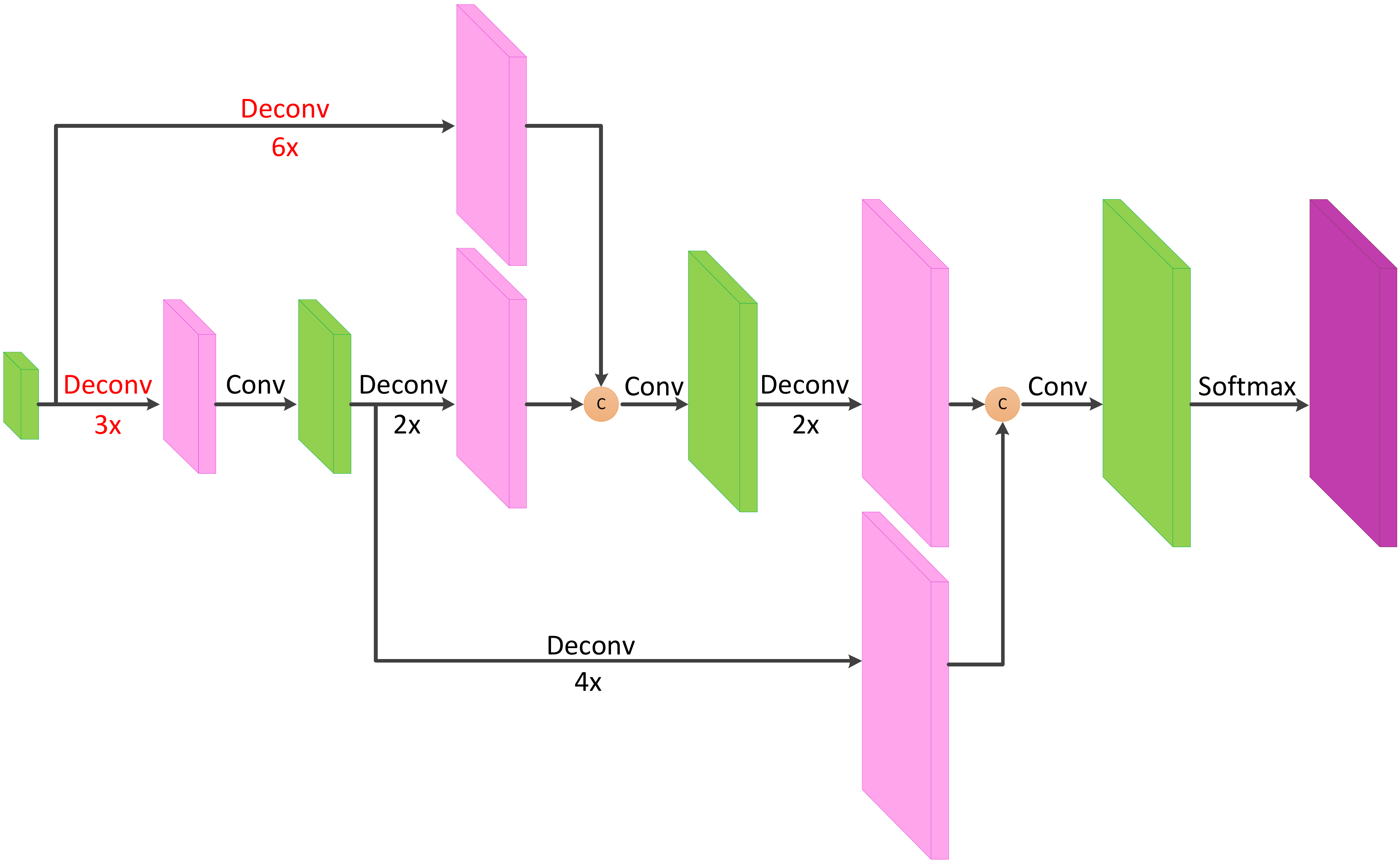} \\
  \end{tabular}
  }
  \caption{Dense connection structure employed in our network. The scale ratios of the first two upsampling layers is 3 and 6 respectively.}
  \label{fig:dense_connection_decoder}
\end{figure}

It is worth mentioning that the number of upsampling paths and the upsampling multiples in the dense connection structure are also variable. The setting of these parameters depends not only on the resolution of encoded feature maps but also on the size of original image. In MS-FCN, since we have downsampled the input image 12 times during the encoding stage, the first two upsampling layers in our model have scale ratios of 3 and 6 respectively as shown in Figure \ref{fig:dense_connection_decoder}.

\section{Experiments and results}

\subsection{Data augmentation}
\label{subsec:decoder}
As part of MICCAI 2009 challenge on automated LV segmentation, the Sunnybrook cardiac dataset contains cine-MRI images with a variety of pathologies and ventricular shapes. This dataset consists of 45 DICOM studies with a resolution of $256 \times 256$ and associated ground truth contours, and it is divided into three parts of 15 cases each: training, validation and online. The training set is used to train our model for LV segmentation, while the validation and online sets are combined as test set to evaluate model performance.

\begin{figure}[!t]
  \centering
  \scalebox{1}{
  \begin{tabular}{c}
    \includegraphics[width=1\linewidth]{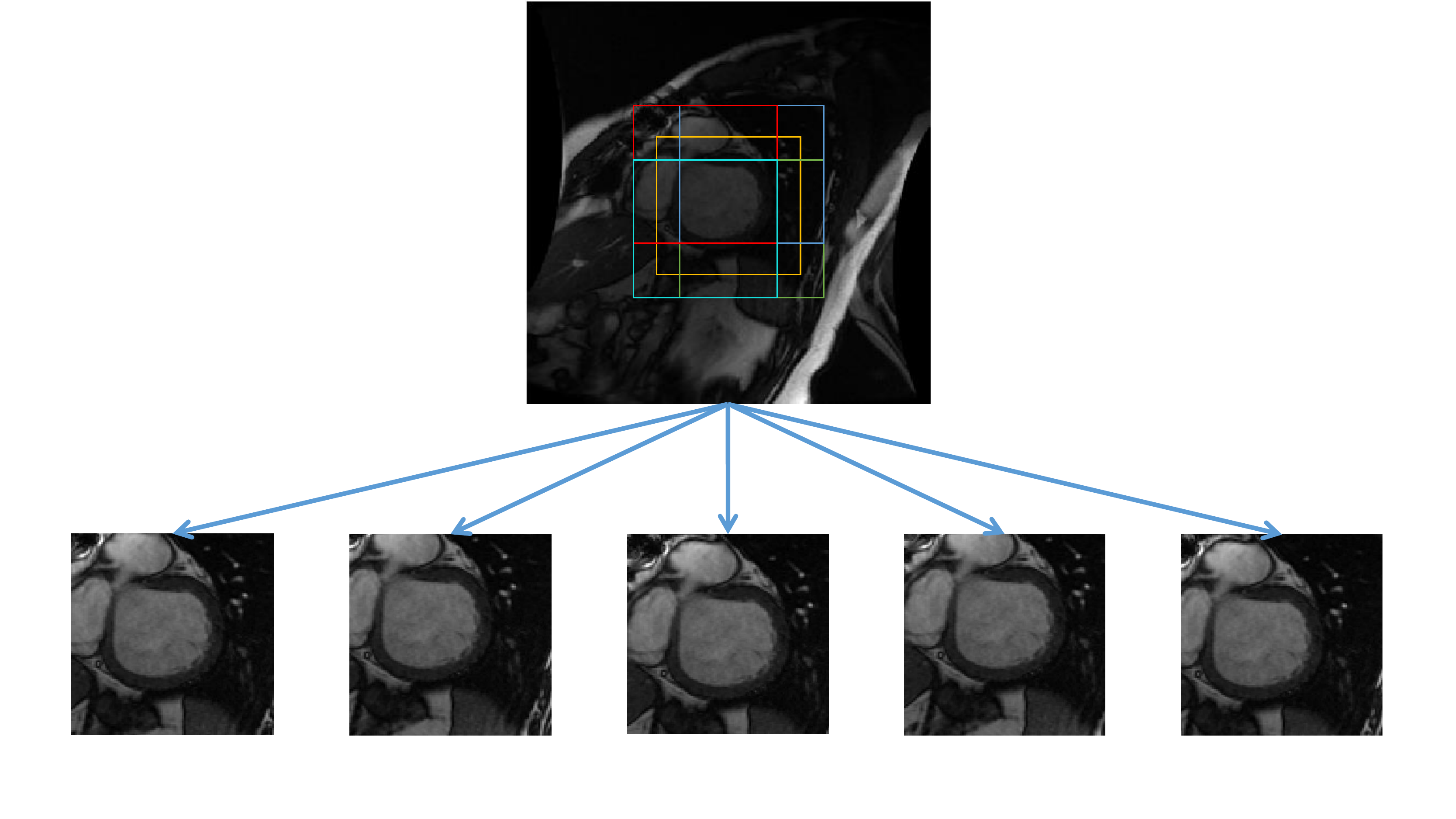} \\
  \end{tabular}
  }
  \caption{Displacement and cropping. Each image is displaced five pixels in four diagonal directions and cropped into $108 \times 108$. }
  \label{fig:displacement}
\end{figure}

Since the amount of Sunnybrook training samples is small, we perform data augmentation, including displacement, rotation, and flipping,  on the training set. First, we use a displacement operation on the training images and each image is translated five pixels in four diagonal directions. Then, we notice that the left ventricle is mostly in the center of image and perform a center cropping. To ensure that all clipped images contain a complete region of interest, we uniformly crop the image to $108 \times 108$ as shown in Figure \ref{fig:displacement}. Meanwhile, the proportion of the left ventricle in the clipped image is relatively large and the class imbalance problem \cite{Buda2017} is relieved to some extent.

\begin{figure}[!t]
  \centering
  \scalebox{1}{
  \begin{tabular}{c}
    \includegraphics[width=1\linewidth]{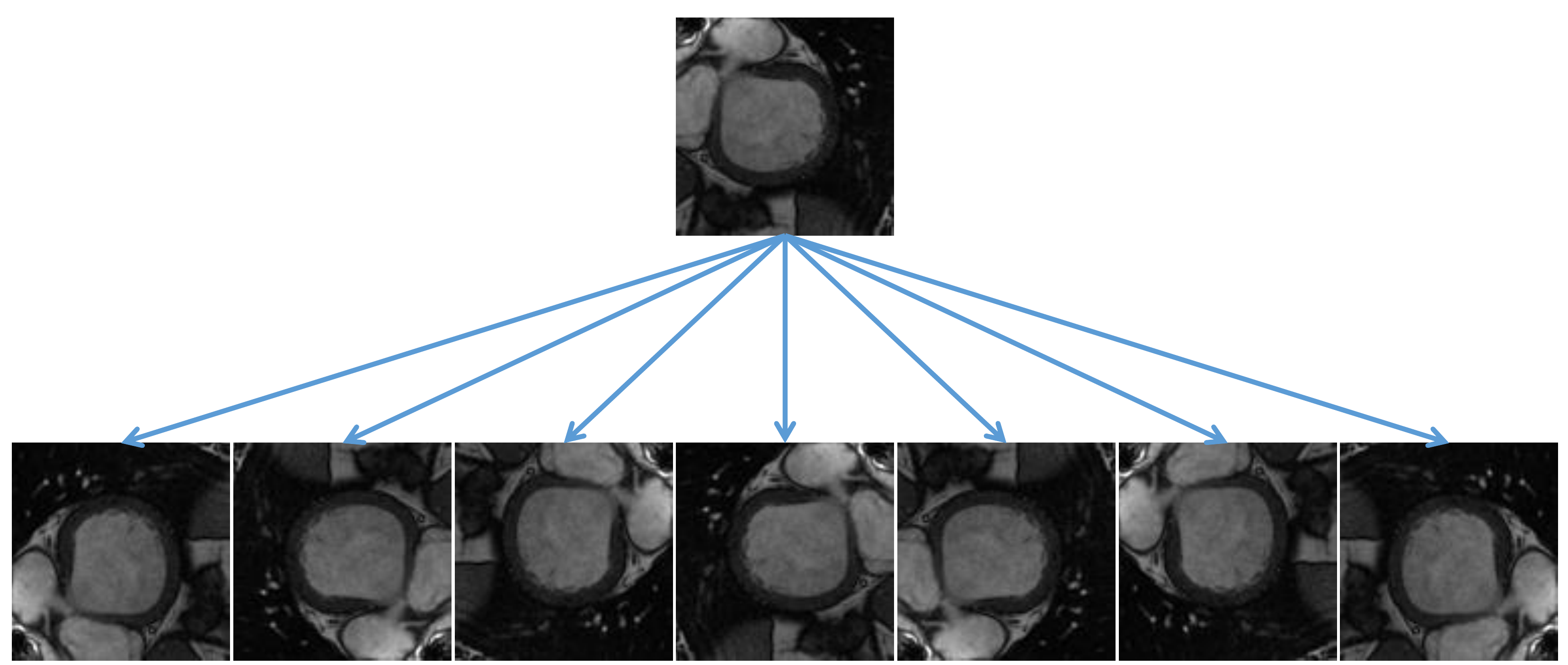} \\
  \end{tabular}
  }
  \caption{Rotation and flipping.}
  \label{fig:rotation}
\end{figure}

Subsequently, we performed rotation and flipping on the cropped images (as shown in Figure \ref{fig:rotation}), including 90 degrees, 180 degrees, and 270 degrees, as well as horizontal flipping and vertical flipping. In conjunction with the displacement operation, the total number of images we get is equal to 40 times that of original images in the training set.

\subsection{Training}
\label{subsec:parameters}
This model is trained and executed with the Caffe \cite{Jia2014} deep learning framework, using a NVIDIA GeForce GTX 1080Ti GPU on Ubuntu 16.04 Linux OS. We employ Nesterov solver with momentum of 0.9 as the optimization algorithm to minimize a cross-entropy cost function calculated from the predicted and the ground truth. We choose the ``Xavier" \cite{Glorot2010} scheme to randomly initialize parameter weights. We also use $L_2$ regularization with a decay coefficient of 0.0005 and a dropout ratio of $p=0.5$ to prevent overfitting of the network. In the training process, we use polynomial decay strategy to reduce the learning rate as follows,

$$base\_lr \times (1 - \frac{iter}{max\_iter})^{power}$$
where $base\_lr = 0.01$ is the initial learning rate, $power = 0.5$ is the decay rate, $iter$ is the current iteration number and $max\_iter$ is the total iteration number which approximately equal to 10 epochs in our experiments.

\subsection{Ablation experiments}
\label{subsec:ms}
In order to verify the efficacy of our network structure mentioned in Section \ref{sec:network}, we have designed a series of ablation experiments as follows. The metrics for assessing the model performance include Dice score, average perpendicular distance (APD) and percentage of good contours \cite{Radau2009}.

\subsection*{Multi-scale pooling module}
\begin{figure}
  \centering
  \subfigure[The first downsampling process implemented by a ordinary pooling layer]{
  \begin{minipage}{1\linewidth}
    \centering
    \includegraphics[width=1\linewidth]{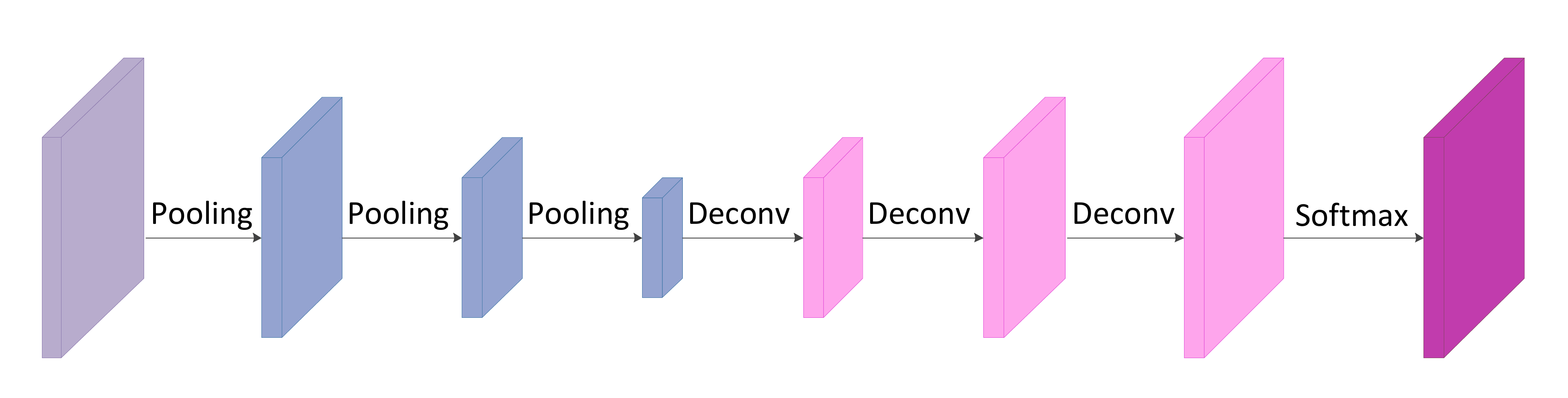}
    \label{fig:without_ms}
  \end{minipage}
  }
  \subfigure[The first downsampling process implemented by the multi-scale pooling module]{
  \begin{minipage}{1\linewidth}
    \centering
    \includegraphics[width=1\linewidth]{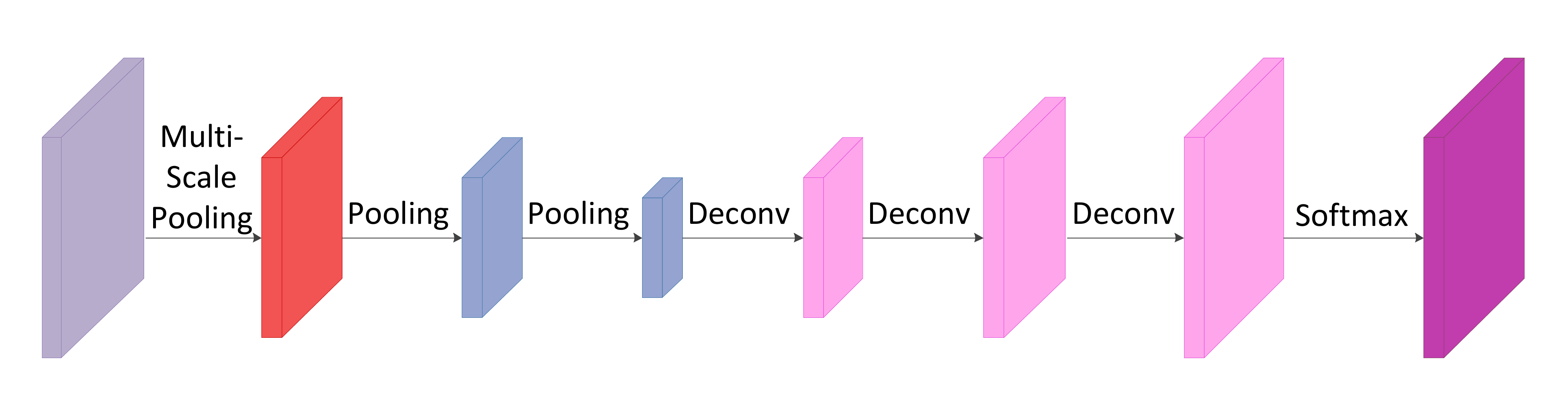}
    \label{fig:with_ms}
  \end{minipage}
  }
  \caption{The first downsampling process implemented by a ordinary pooling layer or the multi-scale pooling module. Convolutional layers are ignored and the multi-scale pooling module is highlighted in red.}
  \label{fig:with_and_without_ms}
\end{figure}

For evaluating the effect of multi-scale pooling module on segmentation performance, we conduct a ablation experiment that whether include this module or not. Figure \ref{fig:with_and_without_ms} shows a simplified diagram of the network structure used in the contrast experiment, ignored the convolutional layers. Figure \ref{fig:with_ms} replaces the first pooling layer in Figure \ref{fig:without_ms} with multi-scale pooling module marked in red.

Table \ref{tab:with_and_without_ms} shows the comparison results. From this table,We can conclude that using the multi-scale pooling module in the encoding stage has better segmentation performance.

\begin{table}[!t]
  \centering
  \caption{Ablation experiment on using a ordinary pooling layer and the multi-scale pooling module as the first downsampling process. The values are provided as mean (standard deviation).}
  \scalebox{1}{
  \begin{tabular}{cccc}
    \toprule
    & & Pooling & Multi-scale\\
    & & layer & pooling module\\
    \midrule
    \multirow{2}*{Dice} & Endo & 0.926(0.023) & \textbf{0.928(0.022)}\\
                        & Epi & 0.954(0.011) & \textbf{0.955(0.011)}\\
    \midrule
    \multirow{2}*{APD(mm)} & Endo & 1.67(0.29) & \textbf{1.61(0.31)}\\
                       & Epi & 1.67(0.26) & \textbf{1.61(0.30)}\\
    \midrule
    Good & Endo & 97.97(5.06) & \textbf{98.35(4.89)}\\
    Contours(\%) & Epi & 96.81(6.26) & \textbf{98.51(3.81)}\\
    \bottomrule
    \end{tabular}
  }
  \label{tab:with_and_without_ms}
\end{table}

\subsection*{Upsampling methods in the multi-scale module}
\label{subsec:upsampling_ms}
Convolutional neural networks generally employ bilinear interpolation or deconvolution as upsampling method. The deconvolution upsampling method uses a deconvolution kernel that can be trained, and theoretically has better learning ability.

In order to evaluate the effects of different upsampling methods in the multi-scale pooling module, we test bilinear interpolation, deconvolution, and group deconvolution, separately, with results listed in Table \ref{tab:upsampling_method}.

\begin{table}[!t]
  \centering
  \caption{Ablation experiment on three different upsampling methods. The values are provided as mean (standard deviation).}
  \scalebox{0.79}{
  \begin{tabular}{ccccc}
    \toprule
    & & Bilinear & Deconvolution & Group\\
    & & interpolation & & deconvolution\\
    \midrule
    \multirow{2}*{Dice} & Endo & 0.924(0.026) & 0.926(0.025) & \textbf{0.928(0.022)}\\
                        & Epi & 0.946(0.017) & 0.949(0.014) & \textbf{0.955(0.011)}\\
    \midrule
    \multirow{2}*{APD(mm)} & Endo & 1.71(0.38) & 1.64(0.33) & \textbf{1.61(0.31)}\\
                       & Epi & 1.82(0.51) & 1.75(0.43) & \textbf{1.61(0.30)}\\
    \midrule
    Good & Endo & \textbf{98.46(3.88)} & 97.69(4.33) & 98.35(4.89)\\
    Contours(\%) & Epi & 97.20(5.49) & 97.09(5.65) & \textbf{98.51(3.81)}\\
    \bottomrule
    \end{tabular}
  }
  \label{tab:upsampling_method}
\end{table}

From the table, we can see that, whether it is for endocardium or epicardium, the use of deconvolution outperform that of bilinear interpolation in the network. For the deconvolution form, using a group deconvolution not only reduces the number of parameters but also has better segmentation performance. Therefore, we use deconvolution upsampling in the multi-scale pooling module, and in order not to increase too many network weights, we set the group parameter to 32.

\subsection*{Dense connection structure}
\label{subsec:dense_connection}
For examining the effect of dense connection structure on the segmentation performance, we trained two separate models with and without dense connection structure. Table \ref{tab:densen_connection} illustrates the comparison results. Compared with the network structure without dense connection structure, the network with dense connection structure improves percentage of good contours and reduces APD significantly, while its dice value of endocardium is only slightly less than the former.

\begin{table}[!t]
  \centering
  \caption{Ablation experiment on whether uses dense connection structure or not in our network. The values are provided as mean (standard deviation).}
  \scalebox{0.9}{
  \begin{tabular}{cccc}
    \toprule
    \multicolumn{2}{c}{Dense connection structure} & Without & With\\
    \midrule
    \multirow{2}*{Dice} & Endo & \textbf{0.928(0.020)} & 0.928(0.022)\\
                        & Epi & 0.954(0.011) & \textbf{0.955(0.011)}\\
    \midrule
    \multirow{2}*{APD(mm)} & Endo & 1.64(0.26) & \textbf{1.61(0.31)}\\
                       & Epi & 1.62(0.29) & \textbf{1.61(0.30)}\\
    \midrule
    Good & Endo & 97.94(4.99) & \textbf{98.35(4.89)}\\
    Contours(\%) & Epi & 98.51(3.88) & \textbf{98.51(3.81)}\\
    \bottomrule
    \end{tabular}
  }
  \label{tab:densen_connection}
\end{table}

\subsection{Comparison and analysis}
\label{subsec:comparison}
In this section, we compare our model with two previous state-of-the-art segment proposal methods: FCN and U-net. It is worth noting that U-net structure performs only three downsampling operations due to the input image resolution is only $108 \times 108$, and it also implements zero pad convolutions to keep image size unchanged. The comparison results are reported in Table \ref{tab:comparison}. As shown in the table, our model has less APD and higher percentage of good contours than FCN and U-net (except the percentage of good endocardial contours). We also achieve a Dice score of $0.93(0.02)$ and $0.96(0.01)$ for endocardium and epicardium respectively, which is the best reported fully automated result on this challenging dataset to date.

\begin{table*}[!t]
  \centering
  \caption{Comparisons with different approaches on Sunnybrook dataset. The values are provided as mean (standard deviation).}
  \scalebox{1}{
  \begin{threeparttable}
  \begin{tabular}{c c c c c c c c}
    \toprule[0.2em]
    \multirow{2}{*}{Method} & \multirow{2}{*}{\#}$^1$ & \multicolumn{2}{c}{Dice} & \multicolumn{2}{c}{APD(mm)} & \multicolumn{2}{c}{Good Contours(\%)}\\
    & & Endo & Epi & Endo & Epi & Endo & Epi \\
    \toprule[0.2em]
    FCN \cite{Tran2016} & 30 & 0.92(0.03) & 0.95(0.02) & 1.74(0.43) & 1.69(0.34) & 97.42(5.88)  & 98.00(3.78) \\
    FCN \cite{Romaguera2017} $^2$ & 45 & 0.92(0.09) & - & - & - & - & -\\
    U-net & 30 & 0.92(0.02) & 0.95(0.01) & 1.71(0.32) & 1.68(0.32) & \textbf{98.66(3.81)} & 96.98(7.01) \\
    MS-FCN & 30 & \textbf{0.93(0.02)} & \textbf{0.96(0.01)} & \textbf{1.61(0.31)} & \textbf{1.61(0.30)} & 98.35(4.89)  & \textbf{98.51(3.81)} \\
    \bottomrule[0.1em]
  \end{tabular}
  \begin{tablenotes}
        \footnotesize
        \item[1] Number of evaluated cases. 30 - validation and online cases, 45 - all cases.
        \item[2] Only provides dice score of endocardial contours.
  \end{tablenotes}
  \end{threeparttable}
  }
  \label{tab:comparison}
\end{table*}

\begin{figure*}[!t]
  \centering
  \scalebox{1}{
  \begin{tabular}{c}
    \includegraphics[width=1\linewidth]{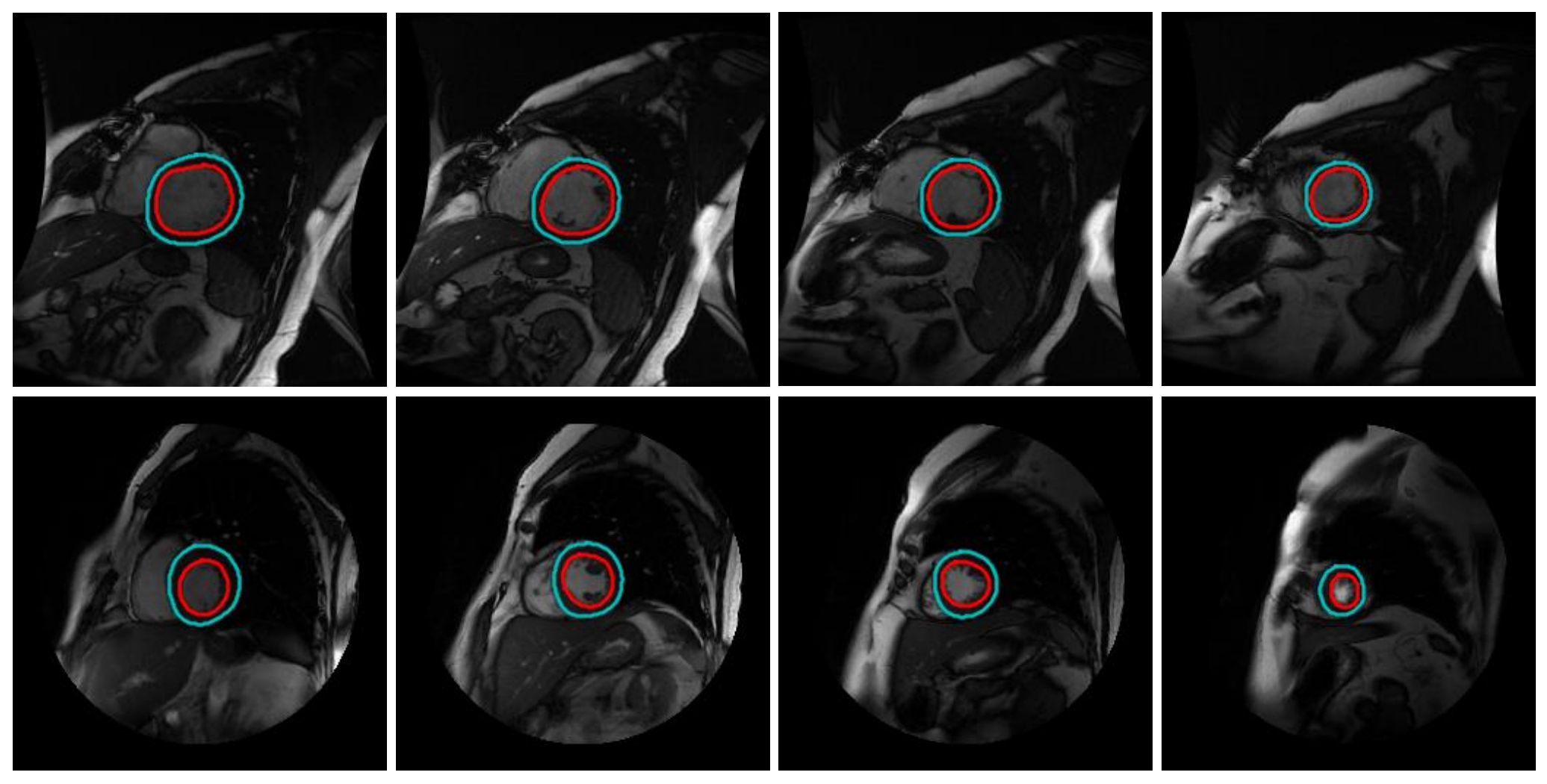} \\
  \end{tabular}
  }
  \caption{Good segmentation examples on Sunnybrook test set. Colors: red - endocardium; cyan - epicardium}
  \label{fig:good_results}
\end{figure*}

\textbf{Good results:} Selected prediction examples from the Sunnybrook test set are shown in Figure \ref{fig:good_results}. As shown in the figure, our model is able to produce accurate results on most slices of left ventricular MRI.

\begin{figure*}[!t]
  \centering
  \scalebox{1}{
  \begin{tabular}{c}
    \includegraphics[width=1\linewidth]{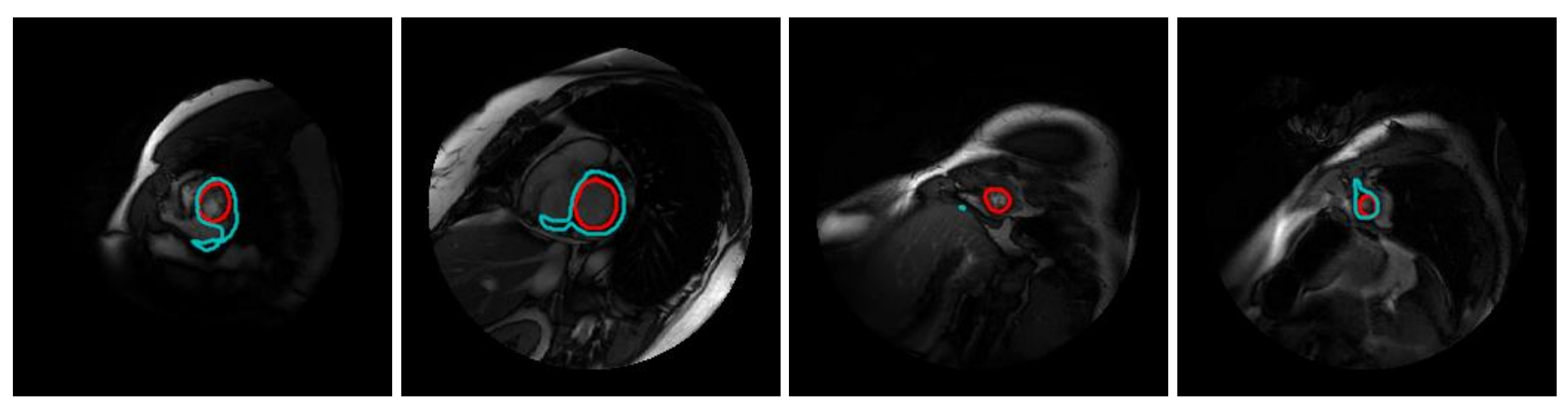} \\
  \end{tabular}
  }
  \caption{Failure segmentation examples on Sunnybrook test set. Colors: red - endocardium; cyan - epicardium}
  \label{fig:failure_results}
\end{figure*}

\textbf{Failure results:} As shown in Figure \ref{fig:failure_results}, our model has difficulty in segmenting (a) left ventricle with ambiguous or imperceptible boundaries, and (b) the apex and basal regions.

\section{Conclusion}
\label{sec:conclusion}
Our proposed model MS-FCN employs the encoder-decoder structure where the multi-scale pooling module is adopted to encode the rich contextual information and the dense connection decoder is used to refine the segmentation results along object boundaries. In order to verify the efficacy of our network structure, we also conduct a series of ablation experiments and compare it with the existing methods. Finally, the experimental results indicate that the proposed model sets a state-of-the-art performance on the Sunnybrook cardiac dataset.

{\small
\bibliographystyle{habbrv}
\bibliography{main}
}

\end{document}